\documentclass[letterpaper]{article} % DO NOT CHANGE THIS
\usepackage{aaai23}  % DO NOT CHANGE THIS
\usepackage{times}  % DO NOT CHANGE THIS
\usepackage{helvet}  % DO NOT CHANGE THIS
\usepackage{courier}  % DO NOT CHANGE THIS
\usepackage[hyphens]{url}  % DO NOT CHANGE THIS
\usepackage{graphicx} % DO NOT CHANGE THIS
\usepackage{natbib}  % DO NOT CHANGE THIS AND DO NOT ADD ANY OPTIONS TO IT
\usepackage{caption} % DO NOT CHANGE THIS AND DO NOT ADD ANY OPTIONS TO IT
\usepackage{algorithm}
\usepackage{algorithmic}

\usepackage{newfloat}
\usepackage{listings}

\usepackage{microtype}
\usepackage{subfigure}
\usepackage{booktabs}
\usepackage{multirow}
\usepackage{amsthm,amsmath,amssymb} 
\usepackage{mathrsfs}
\usepackage{lineno}
\usepackage{xcolor}
\usepackage{color, soul}
\usepackage[inline]{enumitem}
\usepackage{acronym}

\usepackage{bibentry}
\usepackage{natbib}  % DO NOT CHANGE THIS AND DO NOT ADD ANY OPTIONS TO IT
\usepackage{caption} % DO NOT CHANGE THIS AND DO NOT ADD ANY OPTIONS TO IT
\DeclareCaptionStyle{ruled}{labelfont=normalfont,labelsep=colon,strut=off} % DO NOT CHANGE THIS
\floatstyle{ruled}
\newfloat{listing}{tb}{lst}{}
\floatname{listing}{Listing}

\acrodef{NLU}{natural language understanding}
\acrodef{DCT}{debiasing contrastive learning}
\acrodef{NLI}{natural language inference}
\acrodef{ID}{in-distribution}
\acrodef{OOD}{out-of-distribution}
\acrodef{Reweight}{example reweighting}
\acrodef{POE}{product-of-experts}
\acrodef{Conf-reg}{confidence regularization}

\newcommand{\header}[1]{\vspace{1.5mm}\noindent\textbf{#1}.}

\title{Feature-Level Debiased Natural Language Understanding}
\author{
    Yougang Lyu\textsuperscript{\rm 1},
    Piji Li\textsuperscript{\rm 2},
    Yechang Yang\textsuperscript{\rm 1},
    Maarten de Rijke\textsuperscript{\rm 3},
    Pengjie Ren\textsuperscript{\rm 1}\\
    Yukun Zhao\textsuperscript{\rm 1,4},
    Dawei Yin\textsuperscript{\rm 4},
    Zhaochun Ren\textsuperscript{\rm 1}\thanks{Corresponding author.}\\
}
\affiliations{
    \textsuperscript{\rm 1}School of Computer Science and Technology, Shandong University, Qingdao, China\\
    \textsuperscript{\rm 2}College of Computer Science and Technology,
Nanjing University of Aeronautics and Astronautics, Nanjing, China\\
    \textsuperscript{\rm 3}University of Amsterdam, Amsterdam, The Netherlands\\
    \textsuperscript{\rm 4}Baidu Inc., Beijing, China\\
    \{youganglyu, yyc002\}@mail.sdu.edu.cn, pjli@nuaa.edu.cn, m.derijke@uva.nl, jay.ren@outlook.com, zhaoyukun02@baidu.com, yindawei@acm.org, zhaochun.ren@sdu.edu.cn
}

\begin{document}

\maketitle

% !TEX root = ../main.tex

\begin{abstract}
\Ac{NLU} models often rely on \emph{dataset biases} rather than intended task-relevant features to achieve high performance on specific datasets.
As a result, these models perform poorly on datasets outside the training distribution. Some recent studies address this issue by reducing the weights of biased samples during the training process. However, these methods still encode biased latent features in representations and neglect the dynamic nature of bias, which hinders model prediction. 
We propose an \ac{NLU} debiasing method, named \ac{DCT}, to simultaneously alleviate the above problems based on contrastive learning. 
We devise a debiasing, positive sampling strategy to mitigate biased latent features by selecting the least similar biased positive samples. 
We also propose a dynamic negative sampling strategy to capture the dynamic influence of biases by employing a bias-only model to dynamically select the most similar biased negative samples.
We conduct experiments on three NLU benchmark datasets. Experimental results show that \ac{DCT} outperforms state-of-the-art baselines on out-of-distribution datasets while maintaining in-distribution performance. We also verify that \ac{DCT} can reduce biased latent features from the model’s representations.
\end{abstract}

% !TEX root = ../main.tex

\section{Introduction}
\label{sec:introduction}
Pre-trained language models such as BERT \cite{DBLP:conf/naacl/DevlinCLT19} have achieved impressive performance on many \ac{NLU} benchmarks, such as \ac{NLI}~\cite{DBLP:conf/emnlp/BowmanAPM15,DBLP:conf/naacl/WilliamsNB18} and fact verification \cite{DBLP:conf/naacl/ThorneVCM18}. However, recent studies have shown that these models tend to leverage \emph{dataset biases} instead of intended task-relevant features~\cite{DBLP:conf/acl/McCoyPL19,DBLP:conf/emnlp/SchusterSYFSB19,DBLP:journals/corr/abs-2208-11857}.
For example, \citet{DBLP:conf/naacl/GururanganSLSBS18} find that \ac{NLU} models rely on the spurious association between negative words (e.g., \textit{nobody},
\textit{no}, \textit{never} and \textit{nothing}) and \textit{contradiction} labels for prediction in \ac{NLI} datasets, leading to low accuracy on out-of-distribution datasets that lack spurious associations.

To mitigate bias in training datasets, recent \ac{NLU} debiasing methods attempt to train more robust models. Three prevailing debiasing methods exist in \ac{NLU}: 
\begin{enumerate*}[label=(\roman*)]
\item \acf{Reweight} \cite{DBLP:conf/emnlp/SchusterSYFSB19}, \item \acf{POE} \cite{DBLP:conf/emnlp/ClarkYZ19,DBLP:conf/acl-deeplo/HeZW19,DBLP:conf/acl/MahabadiBH20}, and 
\item \acf{Conf-reg} \cite{DBLP:conf/acl/UtamaMG20}.
\end{enumerate*}
These debiasing methods encourage the model to pay less attention to the biased examples, which forces it to learn harder samples to improve \ac{OOD} performance. 

\begin{figure}[t]
  \centering
  \subfigure[Different debiasing methods]{
\centering
\includegraphics[width=0.48\linewidth]{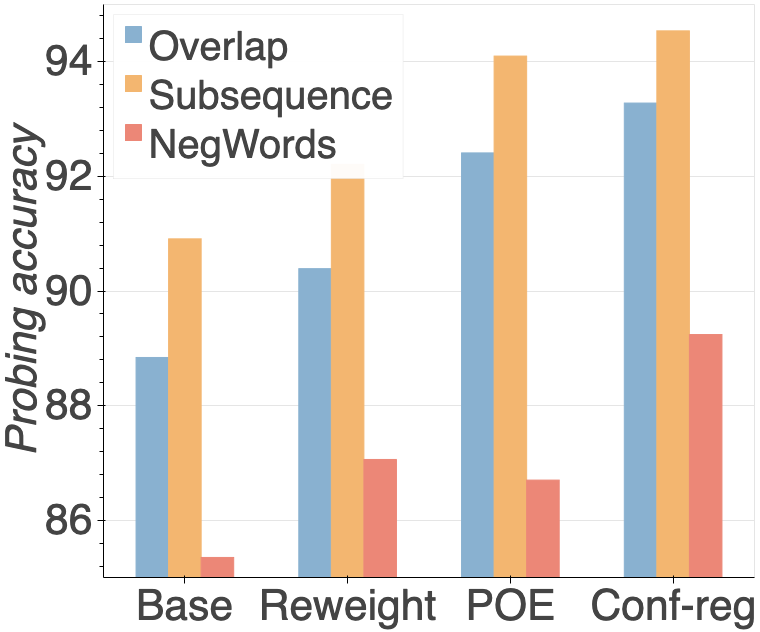}
\label{fig:1a}
}%
\subfigure[Different training epochs]{
\centering
\includegraphics[width=0.48\linewidth]{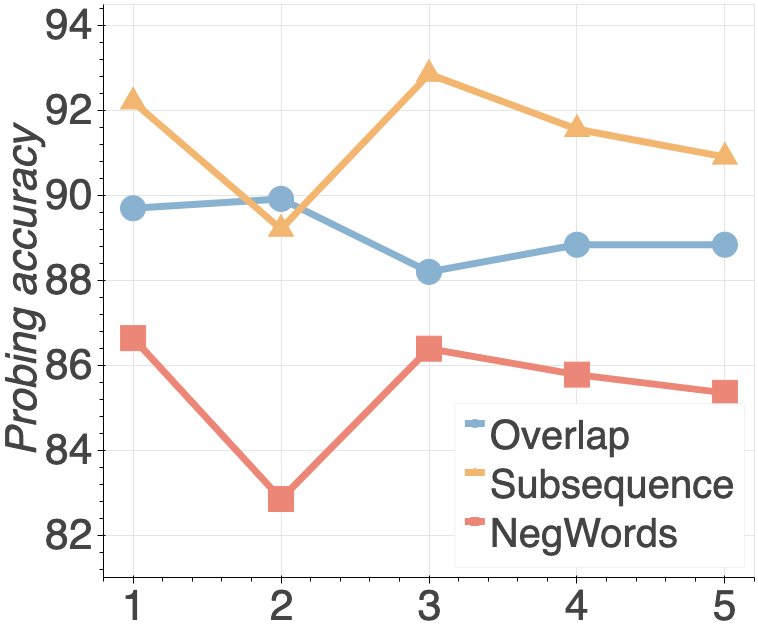}
\label{fig:1b}
}%
\vspace*{-3mm}
\caption{Probing accuracy for three types of biased features (Overlap, Subsequence and Negwords) with different methods and different training epochs on the MNLI dataset. (a) Recent \ac{NLU} debiasing methods (Reweight, POE, and Conf-reg) have a higher probing accuracy of biased features compared to BERT-base. (b) Biased features probing accuracy of BERT-base changes dynamically during the training process.}
\vspace*{-1mm}
\end{figure}

From the perspective of debiasing NLU, two main challenges remain. Both concern \emph{biased features}, that is, features that have spurious correlations with the label, e.g., negative words in input sentences in NLU tasks.
First, existing NLU debiasing methods still encode biased latent features in representations. We follow \citet{DBLP:conf/emnlp/MendelsonB21} to use probing tasks for several types of bias (Overlap, Subsequence and Negwords) to verify whether biased latent features have been removed from  representations.
The probing task for bias is to predict whether a sample is biased based on the model's representation in terms of \emph{probing accuracy}, the accuracy of the probing task. 
Higher probing accuracy of bias means that the model's representation contains more biased features.
Fig.~\ref{fig:1a} shows that existing debiasing methods have a higher probing accuracy of three types of biased latent features from the representations than the fine-tuned BERT-base. 
These results illustrate that existing debiasing methods do not improve \ac{OOD} performance by reducing biased features and modeling intended task-relevant features, but by adjusting the conditional probability of labels given biased features. 
Since these debiasing methods only adjust the conditional probability of labels given biased features, they improve \ac{OOD} performance at the cost of degrading \ac{ID} performance.
This poses a feature-level debiasing challenge to debiasing approaches.

Second, existing NLU debiasing methods neglect the dynamic influence of bias. The number of  biased latent features in a representation changes during the training process. Since the model predicts the label based on the representation, biased features dynamically influence  model prediction during the training process.
In Fig.~\ref{fig:1b} we examine three types of bias and find that the probing accuracy of biases changes during  training. Different types of biased features have a different influence on model prediction during training. It is important to reduce biased features that have the greatest influence on model prediction at different training epochs.
This poses a challenge in capturing the dynamic influence of bias.

To tackle the above challenges, we propose a novel debiasing method, \acfi{DCT}. The main idea of DCT is to encourage positive examples with least similar bias to be closer and negative examples with most similar
bias to be apart at the feature-level. \ac{DCT} consists of two strategies: 
\begin{enumerate*}[label=(\roman*)] 
\item a debiasing, positive sampling strategy to mitigate biased latent features, and 
\item a dynamic negative sampling strategy to capture the dynamic influence of biased features.
\end{enumerate*}
As to the first strategy, \ac{DCT}'s debiasing, positive sampling strategy selects the least similar biased positive samples from the debiasing dataset. We filter debiasing samples from the training set where the bias-only model has high confidence but incorrect predictions. As the bias-only model relies only on biased features to make predictions, the debiasing dataset contains biased features, but the correlation between biased features and labels differs from the dominant spurious correlation in the training dataset. 
As to the second strategy, \ac{DCT}'s dynamic negative sampling strategy uses the bias-only model to dynamically select the most similar biased negative sample during the training process. Furthermore, we adopt \emph{momentum contrast}~\cite{DBLP:conf/cvpr/He0WXG20} to establish a massive queue for saving representations dynamically.
% \begin{table}[tbp]
% \centering
% \caption{Comparison of our method against three prevailing debiasing methods.}
% \label{tab:comparison}
% \resizebox{\linewidth}{!}
% {
% \begin{tabular}{lccc}
% \hline
% Method   & ID acc.      & OOD acc.   & Bias extractability \\ \hline
% RW       & $\Downarrow$ & $\Uparrow$ & $\Uparrow$          \\
% POE      & $\Downarrow$ & $\Uparrow$ & $\Uparrow$          \\
% Conf-reg & $\Uparrow$            & $\Uparrow$ & $\Uparrow$          \\
% DCT (our)      & $\Uparrow$            & $\Uparrow$ & $\Downarrow$    \\ 
% \bottomrule
% \end{tabular}
% }
% \end{table}

We conduct experiments on three \ac{NLU} benchmark datasets to evaluate bias extractability and debiasing performance of our proposed method. DCT outperforms state-of-the-art baselines on \ac{OOD} datasets and maintains \ac{ID} performance by reducing multiple types of biased features in a model's representations.

To sum up, our contributions are as follows:
\begin{itemize}[leftmargin=*,nosep]
    \item To the best of our knowledge, we are the first to focus on feature-level debiasing and modeling the dynamic influence of bias in NLU tasks at the same time.
    \item We propose a novel debiasing method, named DCT, which uses contrastive learning and combines a debiasing, positive sampling strategy and a dynamic negative sampling strategy to reduce biased latent features and capture the dynamic influence of biases.
    \item Experiments on three NLU benchmark datasets show that \ac{DCT} reduces biased latent features in the model's representation and outperforms state-of-the-art baselines on \ac{OOD} datasets while maintaining \ac{ID} performances.\footnote{The code is available at \url{https://github.com/youganglyu/DCT}}
    % $\footnote{Code for reproducing the results in this paper is available at \changed{\url{https://anonymous.4open.science/r/AAAI23_10080}}.}$
\end{itemize}

% !TEX root = ../main.tex

\begin{figure*}[htbp]
  \centering
%   \subfigure[The framework of \acf{DCT}.]{
% \centering
% \includegraphics[width=0.5\textwidth]{AAAI23/figures/framework_ce.pdf}
% \label{fig:2a}
% }%
% \subfigure[Positives and negatives sampling strategies in \ac{DCT}.]{
% \centering
% \includegraphics[width=0.5\textwidth]{AAAI23/figures/framework_dct2.pdf}
% \label{fig:2b}
% }%
\includegraphics[width=1.0\textwidth]{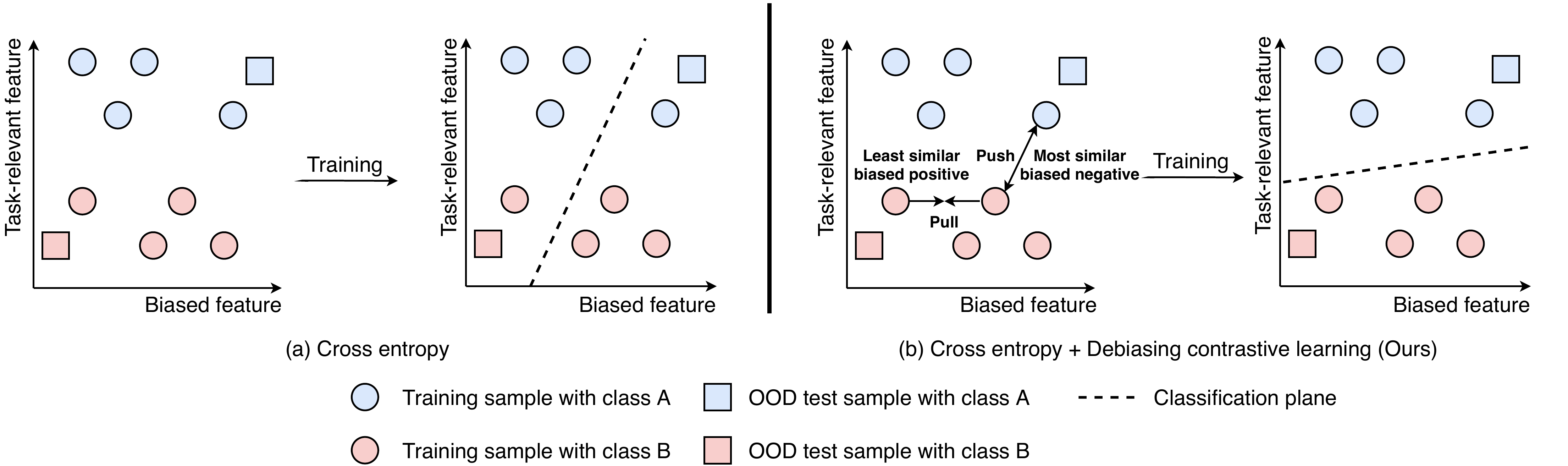}
\vspace*{-5mm}
\caption{(a) Models fine-tuned with cross-entropy use biased features to predict. (b) Models fine-tuned with cross-entropy and \ac{DCT} reduce biased features.} 
\label{fig:2}
\vspace*{-1mm}
\end{figure*}

\section{Related Work}
\label{sec:related_work}

\subsection{Dataset Bias}
Exploiting \emph{dataset biases} seems easier for deep neural networks than learning the intended task-relevant features. \cite{DBLP:journals/natmi/GeirhosJMZBBW20,DBLP:conf/acl/BenderK20}.
For instance, models can perform better than most baselines by only partially using input in \ac{NLI} without capturing the semantic relationships between premises and hypothesis sentences~\cite{DBLP:conf/naacl/GururanganSLSBS18}.
Similar phenomena have been observed in other tasks, e.g., visual question-answering \cite{DBLP:conf/emnlp/AgrawalBP16}, 
reading comprehension \cite{DBLP:conf/emnlp/KaushikL18}, 
and paraphrase identification \cite{DBLP:conf/naacl/ZhangBH19}.
Prior work has constructed challenge datasets consisting of ``counter examples'' to superficial cues that deep neural networks might adopt \cite{DBLP:conf/emnlp/JiaL17,DBLP:conf/acl/GlocknerSG18,DBLP:conf/coling/NaikRSRN18,DBLP:conf/acl/McCoyPL19}.
When models are evaluated on these challenge datasets, their performance often drops to the same as the random baseline~\cite{DBLP:conf/naacl/GururanganSLSBS18,DBLP:conf/emnlp/SchusterSYFSB19}.
Therefore, there is a clear need for methods that are tailored to address NLU dataset biases.

\subsection{Debiasing NLU Methods}
Several studies aim to mitigate dataset bias by improving dataset construction techniques. For example, \citet{DBLP:conf/acl/ZellersHBFC19, DBLP:journals/cacm/SakaguchiBBC21} reduce biased patterns in datasets with adversarial filtering; \citet{DBLP:conf/acl/NieWDBWK20, DBLP:conf/iclr/KaushikHL20} adopt a dynamic, human-in-the-loop data collection technique;  \citet{DBLP:conf/acl/MinMDPL20,DBLP:conf/naacl/SchusterFB21} use adversarial samples to augment the training dataset; and \citet{DBLP:conf/acl/Wu0SD22,DBLP:conf/acl/RossWPPG22} train data generators to generate debiased datasets. 
A complementary line of work trains more robust models with alternative learning algorithms, such as \acf{POE} \citep{DBLP:conf/emnlp/ClarkYZ19,DBLP:conf/acl-deeplo/HeZW19,DBLP:conf/acl/MahabadiBH20}, \acf{Conf-reg} \cite{DBLP:conf/acl/UtamaMG20,DBLP:conf/naacl/DuMJDDGSH21}, and \acf{Reweight} \cite{DBLP:conf/emnlp/SchusterSYFSB19}. 
These algorithms can be formalized as two-stage frameworks; the first step is to train a bias-only model, either automatically \cite{DBLP:conf/emnlp/UtamaMG20,DBLP:conf/iclr/Sanh0BR21,DBLP:conf/acl/GhaddarLRR21} or using prior knowledge about the bias \cite{DBLP:conf/emnlp/ClarkYZ19,DBLP:conf/acl-deeplo/HeZW19,DBLP:conf/starsem/BelinkovPSDR19,DBLP:conf/acl/BelinkovPSDR19}; at the second stage, the output of the bias-only model is used to adjust the loss function of the debiased model. 

However, these debiasing methods make biased features more extractable from the model representations \cite{DBLP:conf/emnlp/MendelsonB21, DBLP:journals/corr/abs-2208-11857}. Instead, we aim to use contrastive learning to dynamically push the NLU model to reduce biased features and capture intended task-relevant features. This enables the NLU model to improve \ac{OOD} performance while maintaining \ac{ID} performance.

\subsection{Contrastive Learning}

The main idea of contrastive learning is to encourage the representation of similar samples to be close and different samples to be apart \cite{DBLP:conf/cvpr/HadsellCL06,DBLP:conf/icml/ChenK0H20}. 
Contrastive learning has been used to improve improve  in-distribution performance in computer vision~\citep[CV;][]{DBLP:conf/cvpr/He0WXG20,DBLP:conf/icml/ChenK0H20} and natural language processing~\cite[NLP;][]{DBLP:conf/acl/GiorgiNWB20,DBLP:conf/acl/WangDLZ20}. 
In a self-supervised framework, positive (i.e., similar) samples can be generated by data augmentation of the anchor sample, and negative (i.e., different) samples can be obtained from the same batch \cite{DBLP:conf/emnlp/GaoYC21} or from a memory bank/queue that saves the representation of previous samples \cite{DBLP:conf/cvpr/WuXYL18,DBLP:conf/icml/ChenK0H20}. In a supervised framework, positive samples belong to the same class while negative samples belong to  a different class \cite{DBLP:conf/nips/KhoslaTWSTIMLK20,DBLP:conf/iclr/GunelDCS21,DBLP:journals/corr/abs-2110-02523}. 

These methods focus on improving \ac{ID} performance, while we aim to use contrastive learning to reduce biased latent features and improve \ac{OOD} performance.

% !TEX root = ../main.tex

% \begin{figure*}[tbhp]
%   \centering
%   \includegraphics[width=0.9\textwidth]{figures/framework4.pdf}
%   \caption{Overview of our proposed DCT.} 
%   \label{fig:framework}
% \end{figure*}

\section{Method}
\label{sec:method}
In this section, we detail the \ac{DCT} method. 
First, we formulate our research problem.
Then, we introduce the overall framework of debiasing contrastive learning. 
Next, we introduce the debiasing, positive sampling strategy and describe the dynamic negative sampling strategy. Finally, a training process with momentum contrast for \ac{DCT} is  explained.

\subsection{Problem Formulation}
\label{ssec:problem_formulatin}
Following~\cite{DBLP:conf/emnlp/ClarkYZ19,DBLP:conf/acl-deeplo/HeZW19,DBLP:conf/acl/MahabadiBH20,DBLP:conf/acl/UtamaMG20}, we formulate \ac{NLU} tasks as a general classification problem. We denote a training dataset as $\mathcal{D}$ consisting of $N$ examples \smash{${\{x_{i},y_{i}\}}_{i=1}^{N}$}, where $x_{i} \in \mathcal{X}$ is the input data, $y_{i} \in \mathcal{Y}$ is the target label, $|\mathcal{Y}|=K$ is the number of the classes.
For each input instance $x$, we assume that the features of $x$ can be divided into intended task-relevant features $x^{t}$ and biased features $x^{b}$, where $x^{t}$ have invariant relations with the label $y$ and $x^{b}$ have spurious relations with the label $y$. The random variables of $x$, $y$, $x^{b}$ and $x^{t}$ are respectively denoted as $X$, $Y$, $X^{b}$ and $X^{t}$.
Our goal is to train a debiasing model $f_{d}$ to capture $\mathbb{P}_{D}(Y|X^{t})$ by reducing biased features $X^{b}$, which performs better on \ac{OOD} datasets and maintains \ac{ID} performance.

\subsection{Debiasing Contrastive Learning}
\label{ssec:dct}
DCT aims to pull the least similar biased positive samples closer to each other and push the most similar biased negative samples apart at the feature-level, as illustrated in Fig.~\ref{fig:2}(b).
To accomplish this, we rewrite the contrastive loss to obtain the \emph{debiasing contrastive learning} loss as follows:
\begin{equation}
\label{eq:dct}
%\resizebox{\linewidth}{!}{
%
\begin{split}
&\mathcal{L}_{DCT}= {}\\
&
\resizebox{.91\linewidth}{!}{
$
\!-\frac{1}{\mid S^{p}_{i}\mid } 
\!\sum\limits_{x_{j}\in S^{p}_{i}}\!\left ( \log\frac{\exp\left (\Phi_{d}(x_{i})\cdot \Phi_{d^{'}}(x_{j})/\tau   \right ) }{\sum\limits_{x_{k}\in S^{n}_{i} / \{x_{j}\}}\exp(\Phi_{d}(x_{i})\cdot \Phi_{d^{'}}(x_{k}) /\tau)  }  \right )\!,\!
$}\!\mbox{}
\end{split}
%$
%}
\end{equation}
where $\Phi_{d}(\cdot)$ refers to the debias encoder, $\Phi_{d}(\cdot)$ denotes to the  momentum encoder, $|S^{p}_{i}|$ is the size of the debiasing, positive sample set $S^{p}_{i}$ for ${x}_{i}$,  $S^{n}_{i}$ denotes the negative sample set for ${x}_{i}$, and $\tau$ is a scalar temperature parameter. 

In the following subsections, we detail the debiasing, positive sampling strategy and the dynamic negative sampling strategy.

\begin{figure}[thbp]
  \centering
  \includegraphics[width=0.9\linewidth]{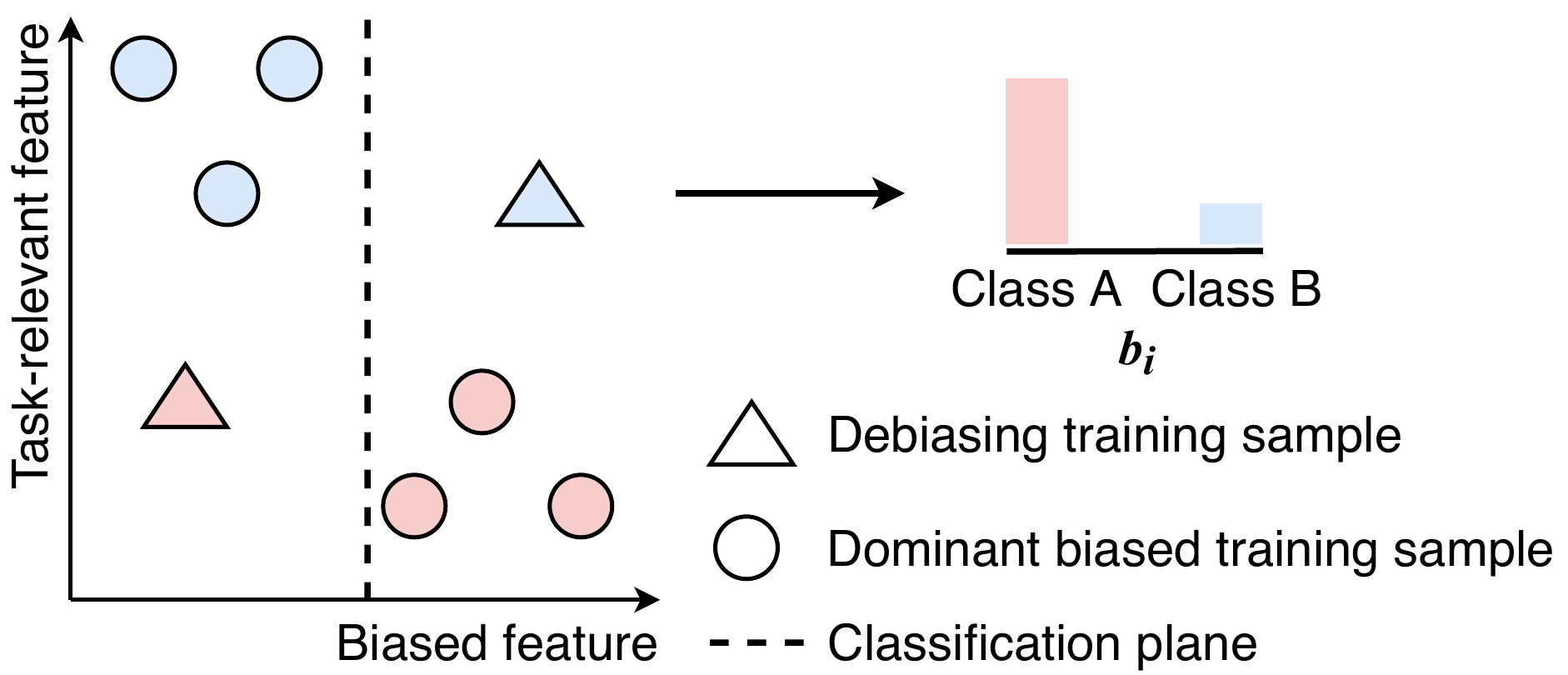}
  \vspace*{-2mm}
  \caption{The bias-only model is used to filter the debiasing dataset from the original training dataset.} 
  \label{fig:dynamic_extrac_acc}
  \vspace*{-1mm}  
\end{figure}

\subsection{Debiasing Positive Sampling}
\label{ssec:debias_pos}
% \todo{
% 1.how to train a bias model \\
% 2.how to choose anti-bias dataset \\
% 3.add anti-bias dataset to dataset \\
% 4.also use anti-bias dataset as dataset for sampling to balancing the feature space \\
% 5.how to construct positive set\\
% 6.add a figure for case\\
% }
% \begin{figure}[htbp]
%   \centering
%   \includegraphics[width=\linewidth]{figures/intra_class2.pdf}
%   \caption{Unbalanced datasets lead to class centers dominated by biased data.}
%   \label{fig:dynamic_extrac_acc}
% \end{figure}
To mitigate biased latent features, we employ a bias-only model to filter the debias dataset from the training dataset and sample least similar biased positive samples from the debias dataset, i.e., we train a bias-only model ${f}_{b}$ to approximate $\mathbb{P}_{D}(Y|X^{b})$ which makes predictions only based on biased features. 
Following \citet{DBLP:conf/iclr/Sanh0BR21}, we use a limited capacity weak learner as the bias-only model, which is trained on the full training dataset. 
As sketched in Fig.~\ref{fig:dynamic_extrac_acc}, if the bias-only model is particularly confident in predicting a sample but predicts incorrectly, it is likely that the sample contains biased features, but the correlation between biased features and labels differs from the dominant spurious correlation in the training dataset. Given a training example $\{x_{i},y_{i}\}$, we assume the output of bias-only model $f_{b}$ to be $b_{i}=\langle b_{i,1},b_{i,2},\ldots,b_{i,K}\rangle$. 
Based on the probabilistic distributions $b_{i}$, we filter the debiasing dataset from the training dataset, so we have:
\begin{equation} 
\label{eq:3}
\mathcal{D}_{debias}=\{\{x_{i},y_{i}\}| b_{i,c}\ge \lambda \wedge y_{i,c}= 0\},
\end{equation}
where $c$ is the predicted class by the bias-only model, $b_{i,c}$ denotes the scalar probability value of the class $c$, $y_{i,c}$ refers to the ground truth of class $c$ for $x_{i}$, and $\lambda$ is a scalar threshold.
% \begin{table}[htbp]
% \label{TAB2}
% % \setlength{\tabcolsep}{15pt}
% \centering
% \caption{Results on HANS.}
% \begin{tabular}{lcc}
% \toprule
% Dataset    & overlap entail    & overlap non-entail \\ \midrule

% MNLI & 99.45 & 25.27 \\
% Anti-bias MNLI  & 1.09 & 98.75 \\
% \bottomrule
% \end{tabular}
% \end{table}

To construct the positive sample set $S^{p}_{i}$ for ${x}_{i}$ in Eq.~\ref{eq:dct}, we select positives that are least similar to ${x}_{i}$ in the debiasing dataset by L2 distance. Additionally, we incorporate the debiasing dataset into the training dataset to generate a sufficient number of positive pairs.

% To construct positive sample set $S^{p}_{i}$ for ${x}_{i}$ in Eq. (\ref{eq:dct}), we combine $\mathcal{S}_{i}^{lp}$ and $\mathcal{S}_{i}^{mp}$ as:
% \begin{align}
% &\mathcal{S}_{i}^{p}=\mathcal{S}_{i}^{mp} \cup \mathcal{S}_{i}^{lp}.
% \end{align}

\subsection{Dynamic Negative Sampling}
\label{ssec:dynamic_neg}
To capture the dynamic influence of bias, we employ the bias-only model to dynamically select the most similar biased negative sample that is closest to the anchor sample ${x}_{i}$. Based on the checkpoints of the bias-only model, we apply the encoder of the bias-only model of each epoch to all the samples and dynamically retrieve the most similar biased negative sample set, that is: 
\begin{equation} 
\mathcal{S}_{i,k}^{dn}=\{x_{j}| y_{j}\ne y_{i} \wedge \operatorname{arg\,min}_{j}(d(\Phi_{b}^{k} (x_{i}),\Phi_{b}^{k}(x_{j})))\},
\end{equation}
where $\mathcal{S}_{i,k}^{dn}$ denotes the set of most similar biased negative samples of $x_{i}$ for the $k$-th epoch, $\Phi_{b}^{k}(\cdot)$ is the bias-only model encoder for the $k$-th epoch, and $d(\cdot)$ refers to the L2 distance function. 
% \begin{figure}[thbp]
%   \centering
%   \includegraphics[width=0.75\linewidth]{figures/dynamic_bias.png}
%   \caption{Extraction accuracy of different types of biased features during the training process.} 
%   \label{fig:dynamic_extrac_acc}
% \end{figure}

To construct the negative sample set $S^{n}_{i}$ for ${x}_{i}$ in Eq.~\ref{eq:dct}, we combine $\mathcal{S}_{i,k}^{dn}$ and other negative samples in the momentum contrast queue.

\subsection{Training with Momentum Contrast}
\label{ssec:training}
To leverage a large number of positive and negative samples, we adopt momentum contrast \cite{DBLP:conf/cvpr/He0WXG20} to build a massive queue for dynamically saving representations.
In order to maintain the representation consistency in the queue, the momentum contrast framework requires two encoders, a debias encoder and a momentum encoder. During training with the momentum contrast framework, the parameter $\theta_{d}$ in  the debias encoder is updated by training samples and then the $\theta_{d^{'}}$ in the momentum encoder is updated by:
\begin{equation} 
\theta_{d^{'}}\gets m\theta_{d}+  \left ( 1-m \right )\theta_{d^{'}}, 
\end{equation}
where $m\in [0,1)$ is a momentum coefficient, which keeps the consistency of sample representations in the queue. The sample representations in the queue are gradually replaced. Specifically, the sample representations encoded by the momentum encoder are added to the queue, and the oldest sample representations are removed. In each training iteration, only the parameter $\theta_{d}$ in the debias encoder is updated by back-propagation.

To directly use the label information, we adopt cross entropy as part of the overall loss for training the main model:
\begin{equation} 
\mathcal{L}_{CE}= - y_{i}\cdot \log f_{d}\left ( x_{i} \right ).
\end{equation}
The overall loss function is formally computed as:
\begin{equation} 
\mathcal{L}= \left ( 1- \alpha  \right ) \mathcal{L}_{CE}+\alpha \mathcal{L}_{DCT},
\end{equation}
where $\alpha $ is a scalar weighting hyperparameter.

\begin{table*}[htbp]
\centering \small
\begin{tabular}{lcc cc cc}
\toprule
& \multicolumn{2}{c}{Overlap} & \multicolumn{2}{c}{Subsequence}  & \multicolumn{2}{c}{NegWords}\\
\cmidrule(r){2-3}
\cmidrule(r){4-5}
\cmidrule(r){6-7}
Method & Compression & Acc.             & Compression               & Acc.            & Compression               & Acc.                       \\ 
\midrule
BERT-base                                          & $3.29 \pm 0.16$        & $ 88.84 \pm 1.22$        & $3.21 \pm 0.24$        & $90.91 \pm 2.53$    & $2.44 \pm 0.12$ & $85.35 \pm 0.93$        \\ 

\midrule
Reweight                                 & $3.68 \pm 0.12$         & $90.39 \pm 0.79$       & $3.66 \pm 0.11$       & $92.21 \pm 2.30 $             & $2.53 \pm 0.11$   & $87.06 \pm 0.22$  \\

POE                                         & $3.68 \pm 0.10$       & $92.41 \pm 1.27$         & $3.82 \pm 0.17 $       & $94.10 \pm 1.97 $           & $2.47  \pm 0.09$  & $86.70 \pm 0.85$      \\

% LMIN                                         & XXX         & XXX         & XXX     & XXX             & XXX  & XXX      \\

Conf-reg                                    & $4.34 \pm 0.20$         & $93.28 \pm 0.69$         & $4.03 \pm 0.18$         & $94.54 \pm 1.82$            &  $2.70 \pm 0.11$ &  $89.24 \pm 0.27$    \\

\textbf{DCT}                                       & $\textbf{2.79} \pm \textbf{0.15}$       & $\textbf{85.35}\pm \textbf{0.84}$                 & $\textbf{2.76}\pm \textbf{0.13}$               & $\textbf{88.96}\pm \textbf{1.77}$             & $\textbf{1.88}\pm \textbf{0.12}$     & $\textbf{80.31}\pm \textbf{0.35}$      \\
\bottomrule
\end{tabular}
\vspace*{-2mm}
\caption{Results of probing for Overlap, Subsequence, and NegWords on MNLI. Acc is the probing accuracy of biases. Note that lower compression scores and probing accuracy represent lower extractability of biased features in the model representation.}
\label{tab:mnli_extract}
\vspace*{-1mm}
\end{table*}

\begin{table*}[htbp]
\centering \small
\begin{tabular}{lcc cc cc}
\toprule
 & \multicolumn{2}{c}{Overlap} & \multicolumn{2}{c}{Subsequence}  & \multicolumn{2}{c}{NegWords}\\
\cmidrule(r){2-3}
\cmidrule(r){4-5}
\cmidrule{6-7}
Method & Compression               & Acc.             & Compression               & Acc.            & Compression               & Acc.                       \\ 

\midrule
BERT-base                                          & $4.39 \pm 0.17$        & $ 93.82 \pm 0.99$        & $5.10 \pm 0.23$        & $94.96 \pm 1.44$    & $4.08 \pm 0.36$ & $93.26 \pm 0.66$        \\ 

\midrule
Reweight                                 & $4.72 \pm 0.28$         & $92.79 \pm 1.05$       & $5.44 \pm 0.50$       & $95.22 \pm 1.62 $             & $4.17 \pm 0.29$   & $93.33 \pm 0.53$  \\

POE                                         & $4.78 \pm 0.14$       & $93.22 \pm 0.47$         & $5.14 \pm 0.15 $       & $94.34 \pm 1.18 $           & $4.28  \pm 0.22$  & $94.12 \pm 0.52$      \\

% LMIN                                         & XXX         & XXX         & XXX     & XXX             & XXX  & XXX      \\

Conf-reg                                    & $5.20 \pm 0.19$         & $94.81 \pm 0.46$         & $5.67 \pm 0.16$         & $93.45 \pm 2.86$            &  $4.78 \pm 0.29$ &  $95.70 \pm 0.49$    \\

\textbf{DCT}                                       & $\textbf{3.14} \pm \textbf{0.10}$       & $\textbf{90.63}\pm \textbf{0.51}$                 & $\textbf{3.44}\pm \textbf{0.15}$               & $\textbf{91.86}\pm \textbf{2.19}$             & $\textbf{2.25}\pm \textbf{0.14}$     & $\textbf{85.17}\pm \textbf{0.75}$      \\
\bottomrule
\end{tabular}
\vspace*{-2mm}
\caption{Results of probing for Overlap, Subsequence and NegWords on SNLI. Notational conventions are the same as in Table~\ref{tab:mnli_extract}.}
\label{tab:snli_extract}
\vspace*{-1mm}
\end{table*}

\begin{table}[htbp]
\centering \small
\begin{tabular}{lcc}
\toprule
Method   & \multicolumn{1}{c}{Compression} & Acc.  \\ \midrule
BERT-base       & $2.57\pm 0.08$                          & $81.82\pm 0.75 $\\ \midrule
Reweight       & $2.68\pm 0.12$                          & $83.72\pm1.88 $\\
POE      & $2.77\pm 0.05 $                          & $84.60\pm1.08 $\\
Conf-reg & $2.82\pm0.06  $                          & $82.47\pm0.90 $\\
\textbf{DCT}      &  $ \textbf{2.21}\pm\textbf{0.08}    $                          &    $ \textbf{78.27}\pm\textbf{1.04}$  \\ \bottomrule
\end{tabular}
\vspace*{-2mm}
\caption{Results of probing for NegWords on FEVER. The notation here is consistent with Table \ref{tab:mnli_extract}.}
\label{tab:fever_extract}
\vspace*{-1mm}
\end{table} 

% !TEX root = ../main.tex

\section{Experiments}
\label{sec:experimental_setup}
\subsection{Research Questions}
We conduct experiments on different NLU tasks to answer the following research questions: 
(RQ1) Does the proposed \ac{DCT} method reduce multiple types of biased features simultaneously? 
(RQ2) How does the proposed \ac{DCT} perform on \ac{ID} and \ac{OOD} datasets compared to state-of-the-art baselines? 
(RQ3) How do strategies and hyperparameters affect \ac{ID} and \ac{OOD} performances of \ac{DCT}? 

\subsection{Datasets}
We use three NLU benchmark datasets in our experiments:
% MNLI, SNLI and FEVER.

\begin{itemize}[leftmargin=*]
\item \textbf{MNLI} -- 
The MNLI dataset \cite{DBLP:conf/naacl/WilliamsNB18}  contains pairs of premise and hypothesis sentences labeled as \textit{entailment}, \textit{neutral}, and \textit{contradiction}. 
We test models trained on MNLI against the challenge dataset HANS \cite{DBLP:conf/acl/McCoyPL19}. 
It contains examples of high overlap between premises and hypothetical sentences but are labeled as \textit{contradiction}. 
Since the overlapping feature is correlated with label \textit{entailment} in MNLI, models trained directly on MNLI tend to perform poorly on HANS.

\item \textbf{SNLI} -- 
The SNLI dataset \cite{DBLP:conf/emnlp/BowmanAPM15} contains pairs of premise and hypothesis sentences labeled as \textit{entailment}, \textit{neutral}, and \textit{contradiction}. Following \citet{DBLP:conf/emnlp/UtamaMSG21}, we evaluate models trained on SNLI against the long and short subsets of the Scramble Test challenge set \cite{DBLP:conf/cogsci/DasguptaGSGG18}. 
It contains samples that are changed the word order against the overlap bias in SNLI dataset.

\item \textbf{FEVER} -- 
The FEVER dataset \cite{DBLP:conf/naacl/ThorneVCM18}  contains pairs of claim and evidence sentences labeled as either \textit{support}, \textit{not-enough-information}, or \textit{refute}. We follow \citet{DBLP:conf/emnlp/SchusterSYFSB19} to process and split the dataset.\footnote{\url{https://github.com/TalSchuster/FeverSymmetric}} FEVER models rely on the claim-only bias, where specific words in the claim are often associated with target label. We evaluate models trained on FEVER against Fever-Symmetric datasets \cite{DBLP:conf/emnlp/SchusterSYFSB19} (version 1 and 2), which were manually constructed to reduce claim-only bias.
\end{itemize}

\subsection{Baselines and Evaluation Metrics}
We compare DCT with three state-of-the-art debiasing methods:
\begin{enumerate*}[label=(\roman*)]
\item \Acf{Reweight} \cite{DBLP:conf/emnlp/SchusterSYFSB19} adjusts the importance of each training instance by computing the importance weight. The weight scalar for each training instance $x_{i}$ is computed as $1-b_{i,g}$, where $b_{i,g}$ is the probability of the bias-only model predicting the gold label. 
\item \Acf{POE} \cite{DBLP:conf/emnlp/ClarkYZ19,DBLP:conf/acl-deeplo/HeZW19,DBLP:conf/acl/MahabadiBH20} trains a debiased model by ensembling with the bias-only model. 
\item \Acf{Conf-reg} \cite{DBLP:conf/acl/UtamaMG20} regularizes model confidence on biased training examples. \ac{Conf-reg} uses a self-distillation training objective and scales the teacher model output by the bias-only model's output.
\end{enumerate*}

To measure the extractability of biased features in the model's representation, we follow \citet{DBLP:conf/emnlp/MendelsonB21} to use compression scores and probing accuracy.\footnote{\url{https://github.com/technion-cs-nlp/bias-probing}} 
The compression score is defined as $compression=\frac{L_{\mathit{unif}}}{L_{\mathit{online}}}$, where $L_{\mathit{unif}}=|D|\log K$ denotes the uniform distribution over
the $K$ labels and $L_{\mathit{online}}$ is the online coding proposed by \citet{DBLP:conf/emnlp/VoitaT20}. The probing accuracy is the accuracy (Acc.) of the probing task.

To evaluate the \ac{ID} and \ac{OOD} performance of models, we follow existing work \cite{DBLP:conf/emnlp/SchusterSYFSB19,DBLP:conf/acl-deeplo/HeZW19,DBLP:conf/emnlp/UtamaMSG21} and employ accuracy (Acc.) on the \ac{ID} and corresponding \ac{OOD} datasets.

\subsection{Implementation Details}
For the MNLI, SNLI and FEVER datasets, we train all models for 5 epochs; all models converge. The base model uses BERT-base \cite{DBLP:conf/naacl/DevlinCLT19} and combines with cross-entropy to fine-tune on three datasets. For debiased models, the first step is to train a bias-only model, where we follow \citet{DBLP:conf/iclr/Sanh0BR21} using TinyBERT \cite{DBLP:conf/emnlp/MichelidF20} for modeling unknown bias. 
Based on the same bias-only model, we train all the above debiased models. In the training process, we adopt the AdamW \cite{DBLP:conf/iclr/LoshchilovH19} optimizer as the optimizer with initial learning rate $3\cdot 10^{-5 }$. Meanwhile, the temperature parameter $\tau$, threshold $\lambda$, momentum coefficient $m$ and scalar weighting hyperparameter $\alpha $ are set to 0.04, 0.6, 0.999, and 0.1. The sizes of the least similar positive samples $S^{p}$ and the most similar negative samples $S^{dn}$ are set to 150 and 1.

% !TEX root = ../main.tex

% !TEX root = ../main.tex

\section{Experimental Results and Analysis}
\label{sec:results}
To answer our research questions we conduct bias extractability experiments, ID and OOD experiments, and ablation studies are conducted.
To directly explore the effectiveness of DCT in reducing biased latent features, we conducted visualization experiments.

\begin{table*}[htbp]
\centering
\small
\begin{tabular}{l cc cc ccc}
\toprule
 & \multicolumn{2}{c}{MNLI (Acc.)} &
\multicolumn{2}{c}{SNLI (Acc.)} &
\multicolumn{3}{c}{FEVER (Acc.)}  \\
\cmidrule(r){2-3}\cmidrule(r){4-5}
\cmidrule{6-8}
Method & dev & HANS & dev & Scramble & dev & Symm. v1 & Symm. v2 \\ 

\midrule
BERT-base                                          & $84.16 \pm 0.23$         & $61.22 \pm 1.17$        & $90.61\pm0.15 $  & $72.74\pm6.87 $ & $87.06 \pm 0.57$     & $56.53 \pm 0.78$     & $63.84 \pm 0.83$          \\ 

\midrule
Reweight                                 & $82.56 \pm 0.31$         & $66.18 \pm 1.04$          & $86.44\pm0.24 $  & $80.30\pm6.99 $ & $83.45 \pm 0.36$     & $61.56 \pm 1.19$             & $67.33 \pm 1.04$      \\

POE                                         & $81.62 \pm 0.18 $        & $67.27 \pm 1.21$          & $83.69\pm0.33 $  & $79.51\pm 6.42$ & $82.23 \pm 0.52$      & $62.19 \pm 1.65$             & $67.36 \pm 1.54$       \\

% LMIN                                         & XXX         & XXX         & XXX     & XXX             & XXX       \\

Conf-reg                                    & $84.15 \pm 0.21$         & $64.89 \pm 1.08$          & $90.56\pm 0.11$  & $83.21\pm4.26 $ & $85.31 \pm 0.29$     &  $59.69 \pm 1.35$             &  $64.75 \pm 1.28$      \\

\textbf{DCT}                                       & $\textbf{84.19} \pm \textbf{0.17}$       & $\textbf{68.30} \pm \textbf{0.85}$   & $\textbf{90.64} \pm \textbf{0.33} $  & $\textbf{86.40}\pm \textbf{4.64}$                & $\textbf{87.12} \pm \textbf{0.34}$               & $\textbf{63.27} \pm \textbf{1.62}$             & $\textbf{68.45} \pm \textbf{1.09}$           \\ 
\bottomrule
\end{tabular}
\vspace*{-2mm}
\caption{Classification accuracy on MNLI, SNLI and FEVER.}
\label{TAB1}
\vspace*{-1mm}
\end{table*}

\subsection{Bias Extractability}
\label{ssec:bias_extract}
For RQ1, we analyze three types of bias and three datasets.
\begin{itemize}[leftmargin=*]
\item \textbf{MNLI} -- Table~\ref{tab:mnli_extract} shows results for the Overlap, Subsequence and Negwords probing tasks on MNLI. Compared to fine-tuned baseline (BERT-base), all debiasing methods except \ac{DCT} increase the extractability of multiple types of biases, which is demonstrated by higher compression values and higher probing accuracy of biases.
Compared to the baselines, \ac{DCT} has the lowest compression value and probing accuracy for multiple biases, indicating that our method \ac{DCT} reduces the extractability of multiple types of biased features simultaneously on MNLI.

\item \textbf{SNLI} -- Table~\ref{tab:snli_extract} shows results for the Overlap, Subsequence and Negwords probing tasks on SNLI. Compared to the fine-tuned baseline (BERT-base), all debiasing methods except \ac{DCT} increase the extractability of multiple biases.
Compared to the baselines, \ac{DCT} has the lowest compression value and bias probing accuracy for multiple biases, indicating that it reduces the extractability of multiple types of biased features simultaneously on SNLI.

\item \textbf{FEVER} -- Table~\ref{tab:fever_extract} shows results for the Negwords probing task on FEVER. Compared to fine-tuned baseline (BERT-base), all debiasing methods except \ac{DCT} increase the extractability of Negwords bias.
Compared to the baselines, \ac{DCT} has the lowest compression value and bias probing accuracy for Negwords bias, indicating that our method \ac{DCT} reduces the extractability of Negwords bias on FEVER.
\end{itemize}

\subsection{ID and OOD performance}
\label{ssec:performance}
Next, we turn to RQ2 and evaluate the in-distribution and out-of-distribution performance of models on the development set and the corresponding challenge set of each dataset.

\header{In-distribution performance} We evaluate the performance of models on the development sets of MNLI, SNLI and FEVER as the \acf{ID} performance. From the \ac{ID} performance of models, we have the following observations:
\begin{enumerate*}[label=(\roman*)]
\item Compared to debiasing baselines, our method \ac{DCT} performs best on the \ac{ID} development sets of MNLI, SNLI and FEVER.
\item Compared to the fine-tuned BERT-base, \ac{Reweight} and \ac{POE} have substantially lower \ac{ID} performance, Conf-reg can maintain \ac{ID} performance on MNLI and SNLI, while \ac{DCT} can maintain \ac{ID} performance on MNLI, SNLI, and FEVER.
\item Although \ac{Conf-reg} also maintains the \ac{ID} performance, it has the highest extractability of biased features. In contrast to Conf-reg, \ac{DCT} has the lowest extractability of biased feature and maintains \ac{ID} performance.
\end{enumerate*}

\header{Out-of-distribution performance}  We evaluate the performance of models on corresponding challenge sets of MNLI, SNLI and FEVER as the \acf{OOD} performance. We observe that: 
\begin{enumerate*}[label=(\roman*)]
\item Compared to the baselines, \ac{DCT} performs well, both on the \ac{ID} dataset and \ac{OOD} dataset. For instance, on the MNLI, \ac{POE} improves the average HANS accuracy from 61.22 to 67.27 but sacrifices 2.54 points of MNLI in-distribution accuracy; \ac{Conf-reg} maintains in-distribution accuracy but only improves 3.67 points on HANS. 
\item Compared to the fine-tuned BERT-base, all debiased baselines improve the \ac{OOD} performance, and \ac{Conf-reg} even achieves a trade-off between \ac{ID} and \ac{OOD} performance, but all debiased baselines improve the extractability of biased features. In contrast, \ac{DCT} reduces the extractability of biased features while improving \ac{ID} and \ac{OOD} performances, indicating that our method \ac{DCT} reduces biased latent features and learns intended task-relevant features.
\end{enumerate*}

\begin{table}[t]
\centering \small
% \resizebox{0.8\linewidth}{!}
% {

\begin{tabular}{lccc}
\toprule
 & \multicolumn{3}{c}{MNLI (Acc.)}   \\
 \cmidrule{2-4}
Method & dev              & HANS            & avg.                   \\ 
\midrule
DCT                                       & 84.19       & \textbf{68.30 }                & \textbf{76.25}          \\ 

-debiasing positive sampling                                      & 83.94       & 65.88                 & 74.91           \\ 

-dynamic negative sampling                                       & 83.54       & 65.79                 & 74.67           \\ 

-all   & \textbf{84.57}       & 63.24                 & 73.91          \\

\bottomrule
\end{tabular}
\vspace*{-2mm}
\caption{Ablation studies with different strategies on MNLI.}
\label{tab:ablation}
\vspace*{-1mm}
\end{table}

% \section{Analysis}
% \label{sec:analysis}
\subsection{Ablation Studies}
\label{ssec:ablation_study}
For RQ3, we perform ablation experiments with respect to strategies, threshold $\lambda$, the number of debiasing positive samples, and the number of dynamic negative samples.

\header{Impact of different strategies} 
Table~\ref{tab:ablation} lists our ablation experiments on MNLI and HANS to explore the effectiveness of strategies. 
\begin{enumerate*}[label=(\roman*)]
\item -debiasing positive sampling: we built the \ac{DCT} model without debiasing dataset and positive samples are randomly sampled from the original training dataset. 
\item -dynamic negative sampling: we built \ac{DCT} without the bias-only model and negative samples are randomly sampled from the original training dataset.
\item -all: we remove the debiasing positive sampling strategy and dynamic negative sampling strategy simultaneously. 
\ac{DCT} is degraded to the original supervised contrastive learning \cite{DBLP:conf/iclr/GunelDCS21}.
\end{enumerate*}

The results in Table~\ref{tab:ablation} show that both strategies (debiasing positive sampling strategy and dynamic negative sampling strategy) enhance the \ac{OOD} performances of \ac{DCT}. 
The \ac{ID} performance will be improved after removing all strategies compared to \ac{DCT}. 
The reason is that supervised contrastive learning aims to improve the \ac{ID} performance. However, the original supervised contrastive learning is not designed for solving dataset bias, so it has the lowest \ac{OOD} performance.

\begin{table}[tbp]
\centering \small
\begin{tabular}{cccc}
\toprule
 & \multicolumn{3}{c}{MNLI (Acc.)} \\
 \cmidrule{2-4}
 Threshold & dev              & HANS            & avg.                   \\ 
\midrule
0.4                                       & 84.12       & 66.27                 & 75.20          \\ 

0.5                                      & 84.15       & 67.13                 & 75.64           \\ 

0.6                                       & 84.19       & \textbf{68.30}                & \textbf{76.25}          \\ 

0.7   & 84.17       & 66.81                 & 75.49          \\
0.8   & \textbf{84.25}       & 64.91                 & 74.58 \\
\bottomrule
\end{tabular}
\vspace*{-2mm}
\caption{Parameter analysis for threshold $\lambda$.}
\label{tab:ablation_threshold}
\vspace*{-1mm}
\end{table}

\header{Impact of threshold $\lambda$}
The threshold $\lambda$ is defined in Eq.~\ref{eq:3} to filter the debiased samples that are incorrectly predicted by the bias-only model but with a high confidence above the threshold.
We conduct an ablation study on the threshold  by changing it
from 0.4 to 0.8. From Table~\ref{tab:ablation_threshold}, when $\lambda$ is set to 0.6, we can get the best \ac{OOD} performance. 
When the value of $\lambda$ is too small, some of the filtered debias samples are not misclassified by containing biased features, thus introducing noise into the debiasing process. 
Conversely, when $\lambda$ is too large, the number and diversity of filtered debias samples are insufficient, thus affecting the debiasing process.
\begin{table}[htbp]
\centering \small
\begin{tabular}{c ccc}
\toprule
 & \multicolumn{3}{c}{MNLI (Acc.)} \\
 \cmidrule{2-4}
$|S^{p}|$                   & dev & HANS & avg. \\ \midrule
\phantom{1,1}75                                     &83.96                         &66.94      &75.45                          \\
\phantom{1,}100                                    &\textbf{84.22}                   &67.15      &75.69                          \\
\phantom{1,}150                                    &84.19                         &\textbf{68.30}      & \textbf{76.25}                          \\
\phantom{1,}500                                    &84.18                         &66.14      &75.16                          \\
1,000                                   &84.12    &66.82      &75.47      \\ \bottomrule
\end{tabular}
\vspace*{-2mm}
\caption{Parameter analysis for the number of debiasing positive samples.}
\label{tab:positive_num}
\vspace*{-1mm}
\end{table}

\header{Impact of the number of debiasing positive samples}
Several experiments are conducted to explore the impact of the number of debiasing positive samples. The results are shown in Table~\ref{tab:positive_num}. More positive samples for \ac{DCT}  do indeed lead to better \ac{OOD} performance. In addition, with the growth
of the number of positive samples, \ac{OOD} performance is slightly
degraded which is probably due to some
positive samples contain similar biased features to the anchor sample.

\begin{table}[htbp]
\centering \small
% \resizebox{0.8\linewidth}{!}
% {
\begin{tabular}{c ccc}
\toprule
& \multicolumn{3}{c}{MNLI (Acc.)}   \\
\cmidrule{2-4}
$|S^{dn}|$ & dev              & HANS            & avg.                   \\ 
\midrule
\phantom{2}0                                       & 83.54       & 65.79                 & 74.67          \\ 

\phantom{2}1                                     & \textbf{84.19}       & \textbf{68.30}                 & \textbf{76.25}          \\ 

\phantom{2}5                                      & 83.66       & 68.25                 & 75.96          \\ 

10   & 84.07       & 68.06                & 76.06          \\ 

20   & 84.11       & 68.15               & 76.13          \\ 

\bottomrule
\end{tabular}
\vspace*{-2mm}
\caption{Parameter analysis for the number of dynamic negative samples.}
\label{tab:negative}
\vspace*{-1mm}
\end{table}

\begin{figure}[thbp]
  \centering
  \subfigure[Encoder of BERT-base.]{
\centering
\includegraphics[width=0.5\linewidth]{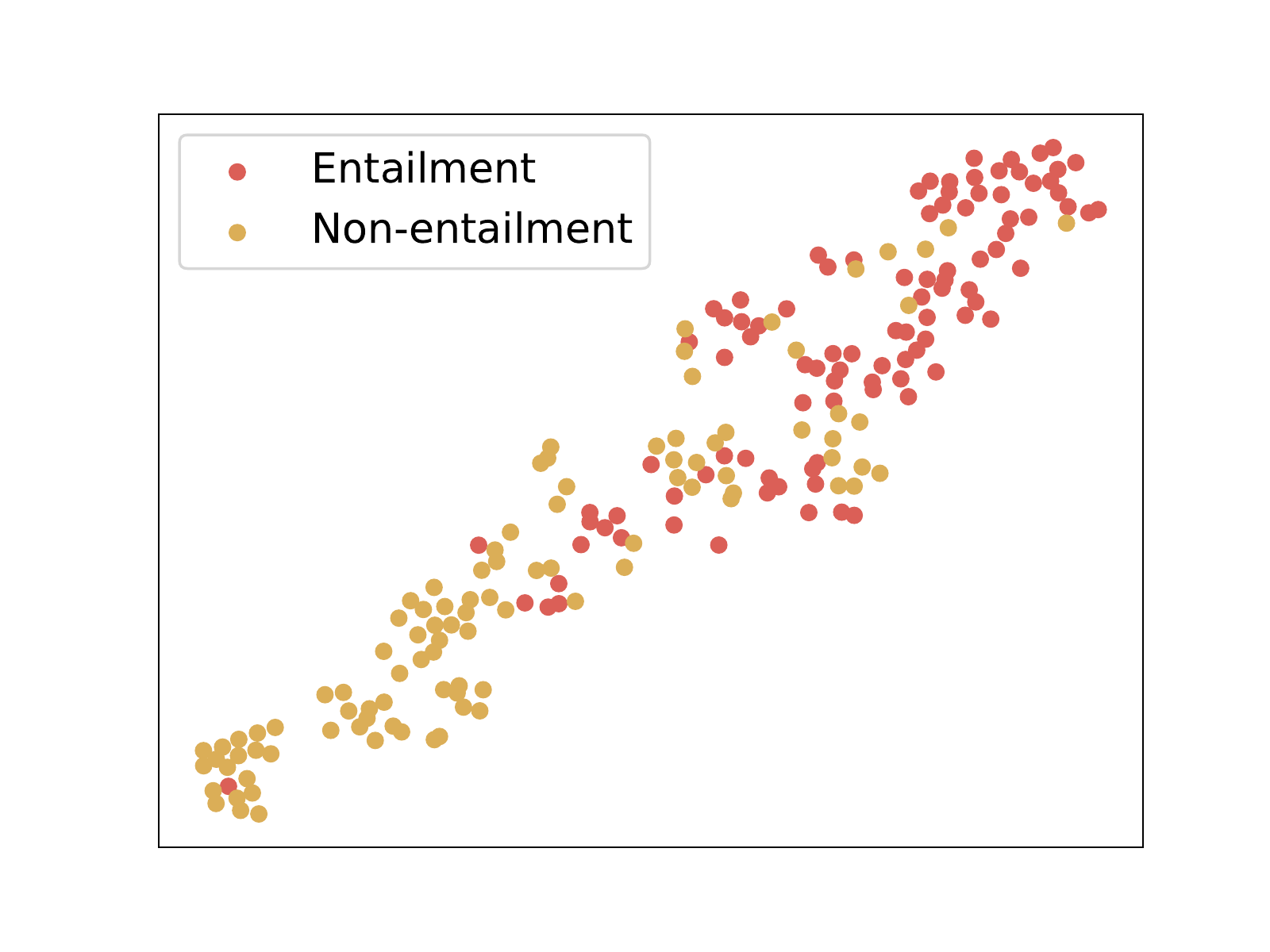}
\label{fig:3a}
}%
\subfigure[Encoder of DCT.]{
\centering
\includegraphics[width=0.5\linewidth]{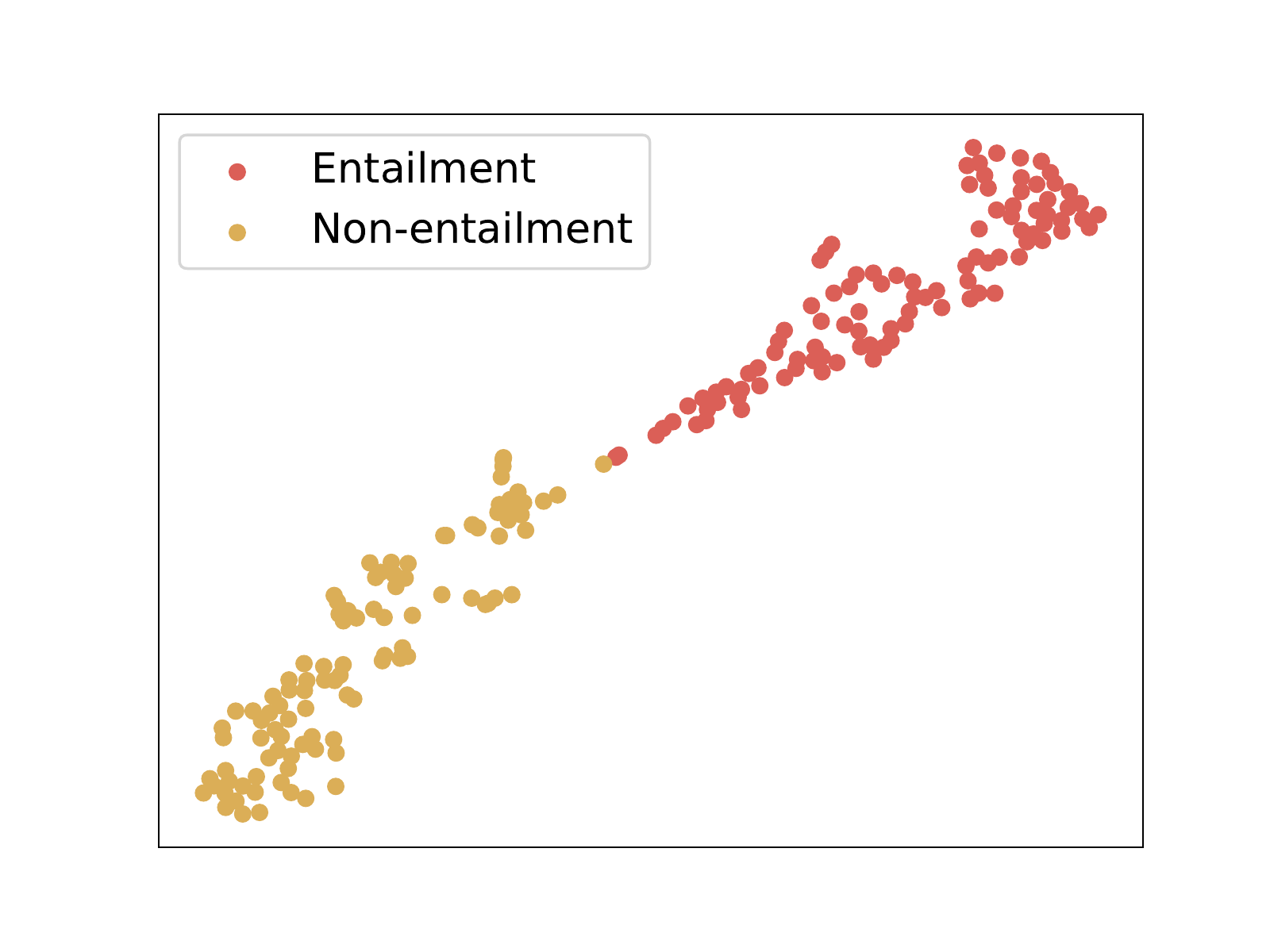}
\label{fig:3b}
}%
\vspace*{-2mm}
\caption{t-SNE plots of the learned [CLS] embeddings on 200 samples from the HANS dataset, comparing BERT-base fine-tuned with cross-entropy only (a) and with our proposed \ac{DCT} (b) for the \ac{NLI} task.
Red: samples contain overlap bias and are labeled as entailment; Yellow: samples contain overlap bias and are labeled as non-entailment.} 
\vspace*{-1mm}
\end{figure}

\header{Impact of the number of dynamic negative samples} 
To explore the impact of the number of dynamic negative samples, sufficient experiments are conducted. As shown in Table~\ref{tab:negative}, the most similar biased negative sample improve the \ac{ID} and \ac{OOD} performances considerably, while adding more similar biased negative samples has little influence on the \ac{ID} and \ac{OOD} performance.
We can observe that the best \ac{ID} and \ac{OOD} performances can be achieved when we set $|S^{dn}|$ to 1.

\subsection{Visualizations}
\label{ssec:visualization}
As mentioned before, the key idea of our proposed method \ac{DCT} is to encourage positive examples with least similar bias to be closer and negative examples with most similar bias to be apart at the feature-level.
To describe this insight intuitively, we use t-SNE to plot the [CLS] representations  of the original BERT finetuned model and our debiased model using 200 data points sampled from the HANS dataset. The 100 data points sampled from HANS containing overlap bias are labeled as entailment and the other 100 data points containing overlap bias are labeled as non-entailment.

As shown in Fig.~\ref{fig:3a}, the encoder trained with only cross-entropy is biased at the feature-level and thus cannot distinguish between samples with the same bias but different classes. In contrast, in Fig.~\ref{fig:3b}, the encoder trained with \ac{DCT} pushes away samples with the same overlap bias at the feature-level, so that samples with the same bias but different classes are more easier to distinguish.

% !TEX root = ../main.tex

\section{Conclusions}
\label{sec:conclusion}
We have focused on reducing biased latent features in an NLU model's representation and on capturing the dynamic influence of biased features. To tackle these challenges, we have proposed an \ac{NLU} debiasing method, namely DCT. To mitigate biased latent features, we have proposed a debiasing, positive sampling strategy. To capture the dynamic influence of biased features, we have devised a dynamic negative sampling strategy to use the bias-only
model to dynamically select the most similar biased negative sample during the training process. 
Experiments have shown that DCT improves the \ac{OOD} performance while maintaining \ac{ID} performance. In addition, our method reduces the extractability of multiple types of bias from an NLU model's representations.
A limitation of DCT is that it is implemented only for the classification task. Our future work is to extend the proposed method to other NLU tasks that are impacted by dataset bias, e.g., named entity recognition and question answering.

\section*{Acknowledgments}
This work was supported by 
the National Key R\&D Program of China with grant No. 2020YFB1406704, 
the Natural Science Foundation of China (62272274, 62202271, 61902219, 61972234, 62072279, 62102234, 62106105), 
the Natural Science Foundation of Shandong Province (ZR2021QF129),
the Key Scientific and Technological Innovation Program of Shandong Province (2019JZZY010129), and 
the Hybrid Intelligence Center, a 10-year program funded by the Dutch Ministry of Education, Culture and Science through the Netherlands Organisation for Scientific Research, \url{https://hybrid-intelligence-centre.nl}.
All content represents the opinion of the authors, which is not necessarily shared or endorsed by their respective employers and/or sponsors.

\bibliography{aaai23.bib}

\end{document}